\documentclass[letterpaper, 10 pt, conference]{ieeeconf}  

\IEEEoverridecommandlockouts                              

\usepackage{graphics} 
\usepackage{epsfig} 
\usepackage{times} 
\usepackage{amsmath} 
\usepackage{amssymb}  
\usepackage{url}  

\usepackage{subcaption}

\usepackage{bbm}
\usepackage[ruled,vlined]{algorithm2e}
\SetKwInput{KwInput}{Input}                
\SetKwInput{KwOutput}{Output}              

\usepackage{booktabs}
\usepackage{multirow}

\title{\LARGE \bf
Guided Uncertainty-Aware Policy Optimization: Combining Learning and Model-Based Strategies for Sample-Efficient Policy Learning}

\author{
Michelle A. Lee$^{*1,2}$, %
Carlos Florensa$^{*1,3}$, %
Jonathan Tremblay$^{1}$, %
Nathan Ratliff$\,^{1}$, \\
Animesh Garg$^{1,4}$, %
Fabio Ramos$^{1,5}$, %
Dieter Fox$^{1,6}$
\thanks{
$^{*}$ Equal Contribution. Work done during an internship at Nvidia.
}
\thanks{
$^{1}$Nvidia, %
$^{2}$Stanford University, %
$^{3}$University of California, Berkeley, %
$^{4}$University of Toronto, %
$^{5}$University of Sydney, %
$^{6}$University of Washington%
}
}

\begin{document}

\makeatletter
\let\@oldmaketitle\@maketitle
\renewcommand{\@maketitle}{\@oldmaketitle
  \includegraphics[width=\linewidth]
    {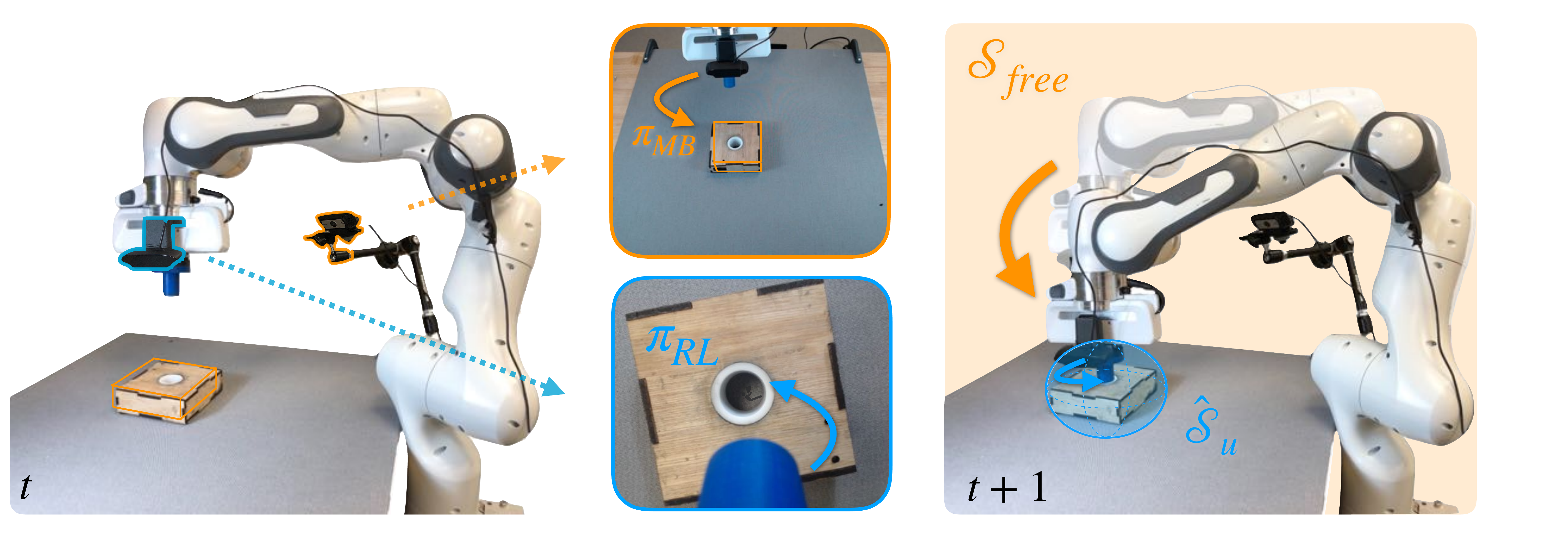} \\[0.25em]
  \refstepcounter{figure}\footnotesize{Fig. 1 Real-world setup for peg insertion: one perception system (in orange) gives the approximate position of the relevant objects. 
  The \textit{model-based} method, $\pi_{\textit{MB}}$, drives the system from free space,
  $\mathcal{S}_{\textit{free}}$, to the uncertainty area, $\hat{\mathcal{S}}_{\textit{u}}$ (in blue). 
  Once inside this area, the model can't be trusted, and a \textit{reinforcement learning} policy, $\pi_{\textit{RL}}$, is then learned directly from the raw sensory inputs of another ``local'' camera (in blue) that gives enough information to complete the task. }
  \label{fig:real} \medskip \vspace{-10pt}}
\makeatother

\maketitle
\thispagestyle{empty}
\pagestyle{empty}


\begin{abstract}

Traditional robotic approaches rely on an accurate model of the environment, a detailed description of how to perform the task, and a robust perception system to keep track of the current state. On the other hand, reinforcement learning approaches can operate directly from raw sensory inputs with only a reward signal to describe the task, but are extremely sample-inefficient and brittle. In this work, we combine the strengths of model-based methods with the flexibility of learning-based methods to obtain a general method that is able to overcome inaccuracies in the robotics perception/actuation pipeline, while requiring minimal interactions with the environment. This is achieved by leveraging uncertainty estimates to divide the space in regions where the given model-based policy is reliable, and regions where it may have flaws or not be well defined. In these uncertain regions, we show that a locally learned-policy can be used directly with raw sensory inputs. We test our algorithm, Guided Uncertainty-Aware Policy Optimization (GUAPO), on a real-world robot performing 
peg insertion. 
Videos are available at: \url{https://sites.google.com/view/guapo-rl}.

\end{abstract}

\section{INTRODUCTION}

Modern robots rely on extensive systems to accomplish a given task, such as a perception module to monitor the state of the world \cite{schmidt2015dart, ChengRMPflowGeneration, xiang2017posecnn}. 
Simple perception failure in this context is catastrophic for the robot, 
since its motion generator relies on it. 
Moreover, classic motion generators are quite rigid in how they accomplish a task, {\em e.g.},
the robot has to pick an object in a specific way and might not recover if the grasp fails. 
These problems make robotics systems unstable, and hard to scale to new domains. 
In order to expand robotics reach we need more robust, adaptive, and flexible systems.

Learning-based method, such as Reinforcement Learning (RL) has the capacity to adapt, and deal directly with raw sensory inputs, which are not subject to estimation errors \cite{nair2018rig, levine2016visumotor}. 
The strength of RL stems from its capacity to define a task at a higher level through a reward function indicating \textit{what} to do, not through an explicit set of control actions describing \textit{how} the task should be performed. 
RL does not need specific physical modelling as they implicitly learn a data-driven model from interacting with the environment \cite{chebotar2018closing}, 
allowing the method to be deployed in different settings.
These characteristics are desired but come with different limitations: 1) randomly interacting with an environment can be quite unsafe for the human users as well as for the equipment,
2) RL is not recognized for being sample efficient. 
As such, introducing RL to a new environment can be time consuming and difficult. 

Classic robotic approaches have mastered generating movements within free space, where there are no contacts with other elements in the environment \cite{ratliff2018rmp}. 
We refer to these accessible methods as Model Based (MB) methods. One of their main limitations is that they normally do not handle perception errors and physical interactions naturally,
{\em e.g.}, grasping an object, placing an object, object insertion, {\em etc.}
As such this limits the expressiveness of roboticists and the reliability of the system. 






In this work we present an algorithmic framework that is aware of its own uncertainty in the perception and actuation system. 
As such a MB guides the agent to the relevant region, hence reducing the area where the RL policy needs to be optimized and making it more invariant to the absolute goal location. Our novel algorithm combines the strengths from MB and RL. 
We leverage the efficiency of MB to move in free-space, and the capacity of RL to learn from its environment from a loosely defined goal. 
In order to efficiently fuse MB and RL, we introduce a perception system that provides uncertainty estimates of the region where contacts might occur. This uncertainty is used to determine the region where the MB method shouldn't be applied, and an RL policy should be learned instead. Therefore, we call our algorithm Guided Uncertainty Aware Policy Optimization (GUAPO).

Figure~\ref{fig:real} shows an overview of our system, the task is initialized with MB where it guides the robot within the 
range of the uncertainties of the object of interest, {\em e.g.}, the box where to insert the peg. 
Once we have reached that region, we switch to RL to complete the task. 
At learning time, we leverage information from task completion by the RL policy to reduce our perception system's uncertainties. 
This work makes the following contributions: 
\begin{itemize}
    \item  We demonstrate that GUAPO outperforms
  pure RL, pure MB, as well as a Residual policy baseline~\cite{silver2018residual, Johannink2019residual} that combines MB and RL for peg insertion; 
    \item We present a simple and yet efficient way to express pose uncertainties for a keypoint based pose estimator; 
    \item We show that our approach is sample efficient for learning methods on real-world robots.
\end{itemize}



\section{DEFINITIONS AND FORMULATION}
We tackle the problem of learning to perform an operation, unknown {\em a-priori}, in an area of which we only have an estimated location and no accurate model. We formalize this problem as a Markov Decision Process (MDP), where we want to find a policy $\pi:\mathcal{S}\times\mathcal{A}\rightarrow \mathbb{R}_+$ that is a probability distribution over actions $a\in\mathcal{A}$ conditioned on a given state $s\in\mathcal{S}$. We optimize this policy to maximize the expected return $\sum_{t=0}^H \gamma^t r(s_t,a_t) $, where $r:\mathcal{S}\rightarrow\mathbb{R}$ is a given reward function, $H$ is the horizon of each rollout, and $\gamma$ is the discount factor.
The first assumption we leverage in this work can be expressed as having a partial knowledge of the transition function $\mathcal{P}:\mathcal{S}\times\mathcal{A}\times\mathcal{S}\rightarrow \mathbb{R}_+$ dictating the probability over next states when executing a certain action in the current state. Specifically, we assume this transition function is available only within a sub-space of the state-space $\mathcal{S}_{\textit{free}}\subset\mathcal{S}$. This is a common case in robotics, where it is perfectly known how the robot moves while it is in free-space, but there are no reliable and accurate models of general contacts and interactions with its surrounding~\cite{shameekcollisions}. 
This partial model can be combined with well established methods able to plan and execute trajectories that traverse $\mathcal{S}_{\textit{free}}$ \cite{RatliffRieMO2015ICRA, SchulmanTrajopt2013, Lav06}, but these methods cannot successfully complete tasks that require acting in $\mathcal{S}_{\textit{u}}=\mathcal{S}\setminus\mathcal{S}_{\textit{free}}$. 
It is usually easy for the user to specify this split relative to the objects in the scene, {\em e.g.}, all points around or in contact with an object are not free space. If we call a point of interest, $g$, in that region we can therefore express that region relative to it as $\mathcal{S}_{\textit{u}}(g)$.  We do not assume perfect knowledge of the absolute coordinates of that point $g$ nor of $\mathcal{S}_{\textit{u}}$, but rather only a noisy estimate of them as described in Section \ref{sec:method_coarse_perception}.

The tasks we consider consist on reaching a particular state or configuration through interaction with the environment, like peg-insertion, switch-toggling, or grasping. These tasks are conveniently defined by a binary reward function $r(s)=\mathbbm{1}[s\in\mathcal{S}_g]$ that indicates having successfully reached a goal set $\mathcal{S}_g\subset\mathcal{S}_{\textit{u}}$, usually described with respect to a point $g$ \cite{Schaul2015uvfa, florensa2018goal, florensa2017reverse}. 
Unfortunately this reward is extremely sparse, and random actions can take an prohibitive amount of samples to discover it \cite{duan2016benchmarking, florensa2017snn}.
Therefore this paper addresses how to leverage the partial model described above to efficiently learn to solve the full task through interactions with the environment, overcoming an imperfect perception system and dynamics.

\section{METHOD}

\begin{figure*}[t!]
\centering
\begin{subfigure}{0.49\textwidth}
    \centering
    \includegraphics[trim={0.25cm 6cm 0.5cm 0cm},clip,width=\linewidth]{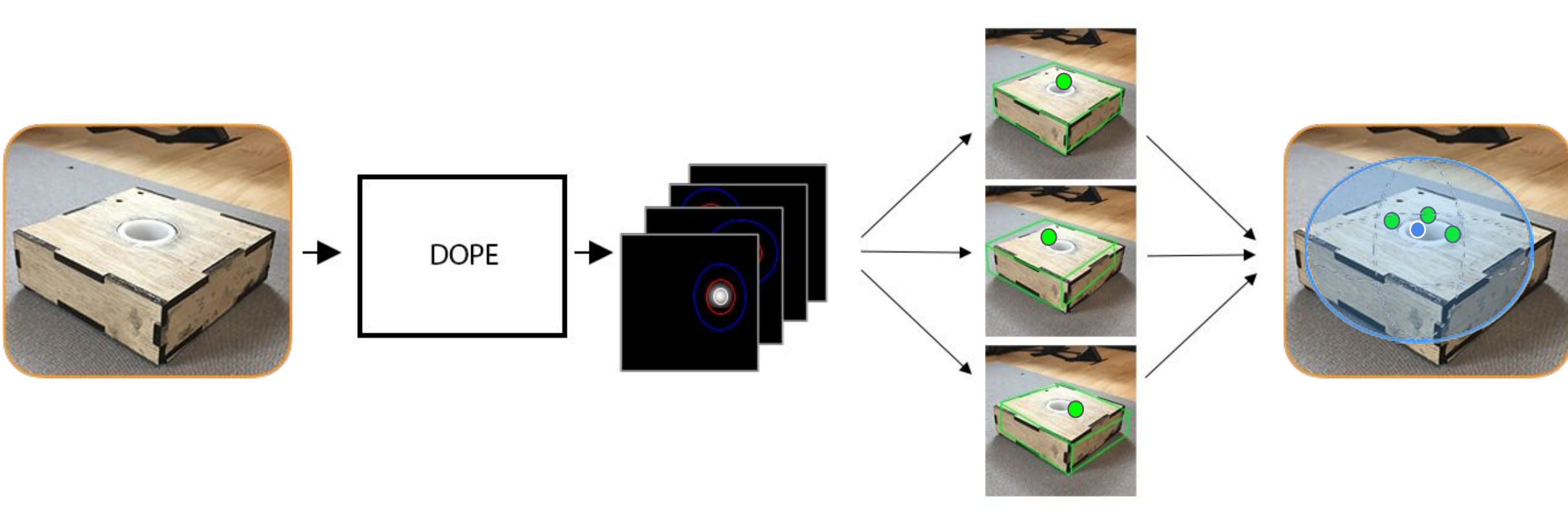}    
    \caption{DOPE perception and uncertainty to estimate $\hat{\mathcal{S}}_{\textit{u}}$}.
    \label{fig:dope}
\end{subfigure}
\hfill
\vline
\begin{subfigure}{0.50\textwidth}
    \centering
    \includegraphics[trim={0.5cm 6cm 0.5cm 0cm },clip,width=\linewidth]{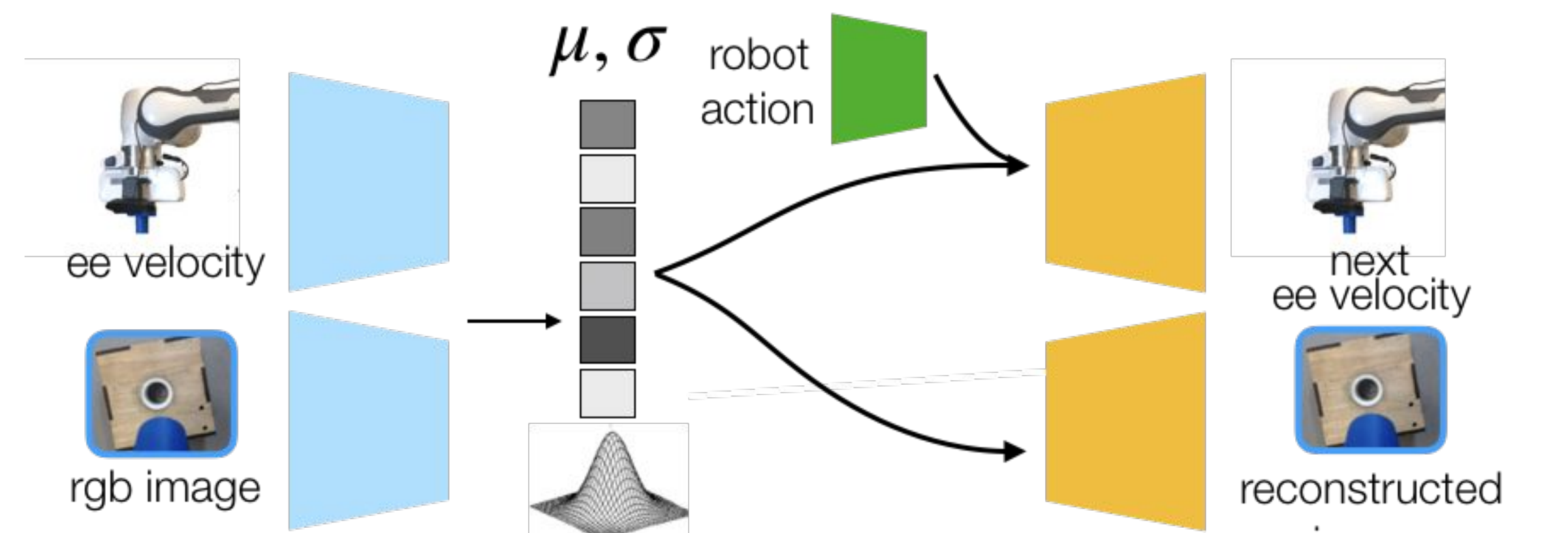}
    \caption{Variational autoencoder for $\pi_{RL}$}.
    \label{fig:vae}
\end{subfigure}
\caption{Perception modules for the model-based component (left) and reinforcement learning component (right).}
\end{figure*}

In this section, we describe the different components of GUAPO. We define $\hat{\mathcal{S}}_{\textit{u}}$ as a super-set of $\mathcal{S}_{\textit{u}}$ generated from the perception system uncertainty estimation. We use this set to partition the space into the regions where the MB method is used, and regions where the RL policy is trained. Then we describe a MB method that can now confidently be used outside of $\hat{\mathcal{S}}_{\textit{u}}$ to bring the robot within that set. Finally we define the RL policy, and how the learning can be more efficient by making its inputs local. We also outline our algorithm in Algorithm~\ref{alg}.

\subsection{From coarse perception to the RL workspace}
\label{sec:method_coarse_perception}

Coarse perception systems are usually cheaper and faster to setup because they might require simpler hardware like RGB cameras, and can be used out-of-the-box without excessive tuning and calibration efforts \cite{tremblay2018deep}. If we use such a system to directly localize $\mathcal{S}_{\textit{u}}$, the perception errors might misleadingly indicate that a certain area belongs to $\mathcal{S}_{\textit{free}}$, hence trying to apply the MB method and potentially not being able to learn how to recover from there. Instead, we propose to use a perception system that also gives an \textit{uncertainty estimate}. Many methods can represent the uncertainty by a nonparametric distribution, with $n$ possible descriptions of the region $\{\mathcal{S}^i_{\textit{u}}\}_{i=1}^n$ and their associated weights $p(\mathcal{S}^i_{\textit{u}})$. By interpreting these weights as the likelihoods $P(\mathcal{S}^i_{\textit{u}}=\mathcal{S}_{\textit{u}})$, we can express the likelihood of a certain state $s$ belonging to $\mathcal{S}_{\textit{u}}$ as:
\begin{equation}
    p(s\in\mathcal{S}_{\textit{u}})=\sum_{i=1}^n \mathbbm{1}\big[s\in\mathcal{S}^i_{\textit{u}}\big] p(\mathcal{S}^i_{\textit{u}}).  \label{eq:S_uncertain}
\end{equation}
If the perception system provides a parametric distribution, the above probability can be computed by integration, or approximated in a way such that the set $\hat{\mathcal{S}}_{\textit{u}}=\{s: p(s\in\mathcal{S}_{\textit{u}})> \epsilon\}$ is a super-set of $\mathcal{S}_{\textit{u}}$ for an appropriate $\epsilon$ set by the user. A more accurate perception system would make $\hat{\mathcal{S}}_{\textit{u}}$ a tighter super-set of $\mathcal{S}_{\textit{u}}$. Now that we have an over-approximation of the area where we cannot use our model-based method, we define a function $\alpha(s)=\mathbbm{1}[s\in\hat{\mathcal{S}}_{\textit{u}}]$ indicating when to apply an \textit{RL} policy $\pi_{\textit{RL}}(a|s)$ instead of the given \textit{MB} one $\pi_{\textit{MB}}$. In short, GUAPO uses a hybrid policy presented in \ref{eq:hard_switch}. 
\begin{equation}
    \pi(a|s) = (1 - \alpha(s))\cdot \pi_{\textit{MB}}(a|s) + \alpha(s)\cdot \pi_{\textit{RL}}(a|s),  \label{eq:hard_switch}
\end{equation}
Therefore we use a switch between these two policies, based on the uncertainty estimate. A lower perception uncertainty reduces the area where the \textit{reinforcement learning} method is required, and improves the overall efficiency. We now detail how each of these policies is obtained. 

\subsection{Model-based actuation}
In the previous section we defined $\hat{\mathcal{S}}_{\textit{u}}$, 
the region containing the goal set $\mathcal{S}_g$ and hence the agent's reward. 
In our problem statement we assume that outside that region, the environment model is well known, and therefore it is amenable to use a \textit{model-based} approach. 
Therefore, whenever we are outside of $\hat{\mathcal{S}}_{\textit{u}}$, the MB approach corrects any deviations. 

Our formulation can be extended for obstacle avoidance. Using a similar approach used to over-estimate the set $\hat{\mathcal{S}}_{\textit{u}}$, we can over-estimate the obstacle set to be avoided by $\hat{\mathcal{S}}^{\textit{obst}}_{\textit{u}}$, and remove that space from where the MB method can be applied, $\mathcal{S}_{\textit{free}}$.
An obstacle-avoiding MB method can be used to get to the area where the goal is, while avoiding the regions where the obstacle might be, as shown in our videos\footnote{https://sites.google.com/view/guapo-rl}.

\subsection{From Model-Based to Reinforcement learning}
Once $\pi_{\textit{MB}}$ has brought the system within $\hat{\mathcal{S}}_{\textit{u}}$, the control is handed-over to $\pi_{\textit{RL}}$ as expressed in Eq.~\ref{eq:hard_switch}. Note that our switching definition goes both ways, and therefore if $\pi_{\textit{RL}}$ takes exploratory actions that move it outside of $\hat{\mathcal{S}}_{\textit{u}}$, the MB method will act again to funnel the state to the area of interest. 
This also provides a framework for safe learning \cite{fisac2017safe_learning} in case there are obstacles to avoid as introduced in the section above. There are several advantages to having a more restricted area where the RL policy needs to learn how to act: first the exploration becomes easier, second, the policy can be local. In particular, we only feed to $\pi_{\textit{RL}}$ the images from a wrist-mounted camera and its current velocities, as depicted in Fig.~\ref{fig:vae}. Not using global information from our perception system in Fig.~\ref{fig:dope} can make our RL policy generalize better across locations of $\hat{\mathcal{S}}_{\textit{u}}$. Finally, we propose to use an off-policy RL algorithm, so all the observed transitions can be added in the replay buffer, no matter if they come from $\pi_{\textit{MB}}$ or $\pi_{\textit{RL}}$.

\subsection{Closing the MB-RL  loop}
This framework also allows to use any newly acquired experience to reduce $\hat{\mathcal{S}}_{\textit{u}}$ such that successive rollouts can use the \textit{model-based} method in larger areas of the state-space. 
For example, in the peg-insertion task, once the reward of fully inserting the peg is received, the location of the opening is immediately known. Since we no longer need to rely on the noisy perception system to estimate the location of the hole, we can update $\hat{\mathcal{S}}_{\textit{u}}=\mathcal{S}_{\textit{u}}$, where now the reinforcement learning algorithm only needs to do the actual insertion and not keep looking for the opening.

\begin{algorithm}
\SetAlgoLined
\KwInput{
$s_0 \in \mathcal{S}$: reset state 

$o^{g}$: global observation $\rightarrow$ RGB workspace camera

$o^{l}$: local observations $\rightarrow$ wrist-mounted  camera, robot velocity observations

$\mathcal{S}_{u}$: model uncertain region containing the goal region $\mathcal{S}_{g}$, both unknown a-priori until reward is obtained.
}   

\KwOutput{$a_t$: robot actions } 

\hrulefill

\textit{goal\_localized} $\leftarrow$ False 

\For{each episode} { 

$s_{t=0}\leftarrow s_0$ \tcp*{Reset robot}

\If{\textbf{not} goal\_localized} {

$\hat{\mathcal{S}}_{u} \leftarrow $ DOPE($o^{g}$) \tcp*{Run perception}
}

    \For{$t=0\dots H$}{
    
        \If{robot \textbf{not} in $\hat{\mathcal{S}}_{\textit{u}}$} {
        
            $a_t \leftarrow \pi_{\textit{MB}}$ \tcp*{Takes robot to $\hat{\mathcal{S}}_{u}$}
        }
        \Else {
         $a_t \leftarrow \pi_{\textit{RL}}(o^{l}_t)$ ;
        }
        
        Apply action $a_t$ to environment ;
        
        $r_t=\mathbbm{1}[s_t \in\mathcal{S}_g]$
        
        Add $(o^l_t, a_t, o^l_{t+1}, r_t)$ to replay buffer ;
        
        \If{$r_t$ = 1} {
            $\hat{\mathcal{S}}_{u} \leftarrow \mathcal{S}_{u}(s_t)$ \tcp*{No DOPE uncert.}
            
            \textit{goal\_localized} $\leftarrow$ True ;
        }
    } 

}

 \caption{GUAPO}
 \label{alg}
\end{algorithm}

\section{IMPLEMENTATION DETAILS}
Here we describe the implementation details of our GUAPO algorithm for a peg insertion task with a Franka Panda robot (7-DoF torque-controlled robot). We first introduce the perception module and how an uncertainty estimate is obtained to localize $\hat{\mathcal{S}}_{\textit{u}}$. Then we describe the model-based policy used to navigate in $\mathcal{S}_{\textit{free}}$ while avoiding obstacles, and the RL algorithm and the architecture of the RL policy being learned. Finally, we introduce our task set-up, the baseline algorithms we compare GUAPO with, and their implementations. 

\begin{figure}
    \centering
    \begin{tabular}{cc}
        \includegraphics[width=0.45\linewidth]{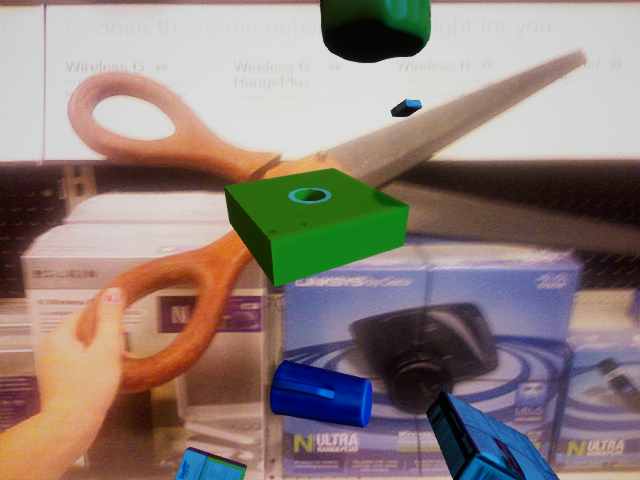} &
        \includegraphics[width=0.45\linewidth]{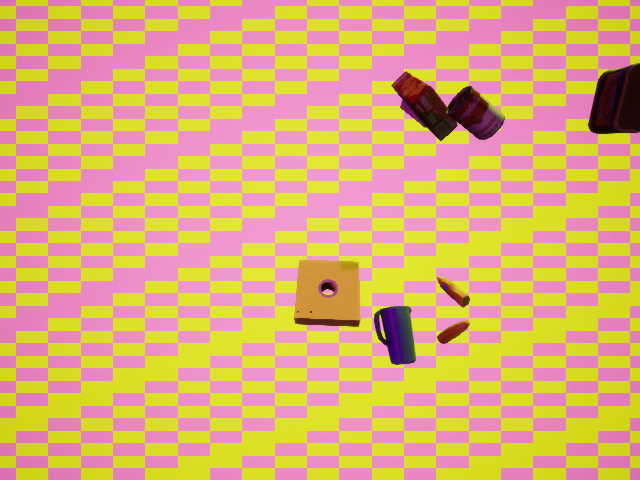} \\
    \end{tabular}
    \caption{Representative synthetic training images for our hole box.}
    \vspace{-0.7cm}
    \label{fig:dope_dr}
\end{figure}

\subsection{Perception Module}
We use Deep Object Pose Estimator (DOPE)~\cite{tremblay2018deep} as the base for our perception system. DOPE uses a simple neural network architecture that can be quickly trained with synthetic data and domain randomization using NDDS \cite{to2018ndds}. 
Figure~\ref{fig:dope_dr} shows generated images with domain randomization used to train our perception system and thus allowing domain transfer (from synthetic to real world). 
Note that the model of the object that DOPE needs to detect is not very detailed, consisting of the approximate shape without texture.
This is a challenging case, specially because no depth sensing is used to supplement the RGB information.
DOPE algorithm first finds the object cuboid keypoints using local peaks on the map. 
Using the cuboid real dimensions, camera intrinsics, and the keypoint locations, DOPE runs a PnP algorithm \cite{pnp_fua} to find the final object pose in the camera frame. 
 
For this work we extended the DOPE perception system to obtain uncertainty estimates of the object pose. 
This extension augments the peak estimation algorithm by fitting a 2d Gaussian around each found peak, as depicted by the dark contour maps in Fig.~\ref{fig:dope}.
We then run PnP algorithm on $n$ set of keypoints, where each set of keypoint is constructed by sampling from all the 2d Gaussians. This provides $n$ possible poses of the object consistent with the detection algorithm, as drawn in green bounding boxes in Fig.~\ref{fig:dope}. In this work we treat them as equally likely.

From our problem formulation, we assume access to a rough description of the area of interest, $\mathcal{S}_{\textit{u}}$, around the object where an operation needs to be performed. In our peg insertion task, this is a rectangle centered at the opening of the hole. For each of the $n$ pose samples given by our extended DOPE perception algorithm, we compute the associated hole opening positions, $\{h_i\}_{i=1}^n$, represented by the green dots in Fig.~\ref{fig:dope}. These points are then fitted by 3d Gaussian with diagonal covariance, represented in blue in the same figure. We use the mean $\hat{\mu}_{\textit{hole}}=\frac{1}{n}\sum_{i=1}^n h_i$ as the center of $\hat{\mathcal{S}}_{\textit{u}}$ and we over-approximate Eqn.~\ref{eq:S_uncertain} by displacing $\mathcal{S}_{\textit{u}}$ along the axis by one standard deviation.

The perception module setup is depicted in Fig.~\ref{fig:real} in orange, where the camera for DOPE (640x480x3 RGB images from Logitech Carl Zeiss Tessar) is mounted overlooking our workspace. The top center image with the orange border is a sample from that camera.

\subsection{Model-Based Controller Design}
As model-based controller, we use a target attractors defined by Riemannian Motion Policies (RMPs) \cite{ratliff2018rmp} to move the robot towards a desired end-effector location. The RMPs take in a desired end-effector position $\mathbf{x} \in \mathbb{R}^3 $ in Cartesian space. The target is set to be the centroid of $\hat{\mathcal{S}}_{\textit{u}}$, which in our case corresponds to the opening of the hole $\hat{\mu}_{\textit{hole}}$. As explained in the previous section, a coarse model of the object is required to train a perception module able to provide this location estimate and its uncertainty. The RMPs also require a model of the robot. These two requirements are the ones that give this part of the method the ``model-based" component. By utilizing the RL component described in the following section, our GUAPO algorithm does not need these models to be extremely accurate.
In case obstacles need to be avoided to reach $\hat{\mathcal{S}}_{u}$, we can define barrier-type RMPs. 

The policies are sending end-effector position commands at 20 {\em Hz}. 
The RMPs are computing desired joint positions $\mathbf{q}_d$ at 1000 {\em Hz}. Given that impedance-end-effector control is an action space which has been shown to improve sample efficiency for policy learning for RL \cite{martin2019variable}, we also use the RMPs interface as our reinforcement learning action space.

\subsection{Reinforcement Learning Algorithm and Architecture}
We use a state-of-the-art model-free off-policy RL algorithm, Soft Actor Critic~\cite{Haarnoja2018sac}.
The RL policy acts directly from raw sensory inputs. This consists on joint velocities and images from a wrist-mounted camera (64x64x3 RGB images from 
a Logitech Carl Zeiss Tessar) on the robot (see Fig.~\ref{fig:real}). As illustrated in Fig.~\ref{fig:vae}, all inputs are fed into a $\beta$-VAE \cite{higgins2017betavae}. The VAE gives us a low-dimensional latent-space representation of the state, which has been shown to improve sample efficiency of RL algorithms~\cite{StateReprLearning}. The parameters of this VAE are trained before-hand on a data-set collected off-line. The only part that is learned by the RL algorithm is a 2-layer MLP that takes as input the 64-dimensional latent representation given by the VAE, and produces 3D position displacement $\Delta \mathbf{x}$ of the robot end-effector.

\subsection{Training Details}
The VAE is pre-trained with 160,000 datapoints for 12 epochs, on the Titan XP GPU. DOPE is trained for 8 hours on 4 p100 GPU. All our learning-based policy methods (GUAPO, SAC baseline, and the Residual Policy baseline described in Sec.~\ref{sec:experimental}) were trained for 60 training iterations. In total, each policy was trained with 120 training episodes, as each iteration has two training episodes, each with 1000 steps. This takes 90 min. to train.

\begin{figure*}[ht!]
\centering
\includegraphics[trim={0cm 2.75cm 0cm 0.7cm},clip,width=.85\linewidth]{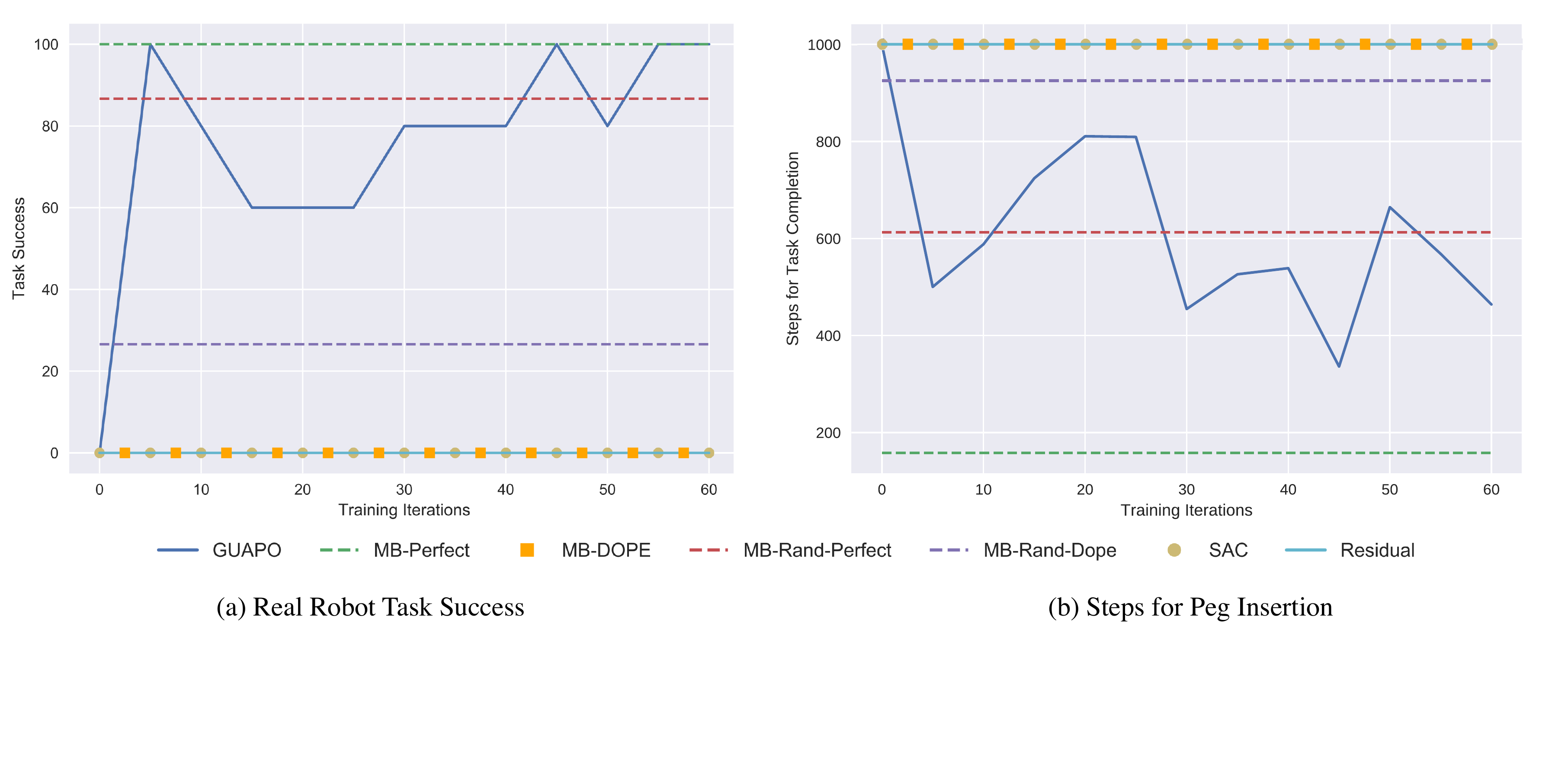}
\caption{GUAPO is compared with five other methods: (1) Model-based policy with perfect goal estimate (MB-Perfect), (2) Model-based policy with additive random actions and perfect goal estimate (MB-Rand-Perfect), (3) Model-Based policy with additive random actions using DOPE goal estimates (MB-Rand-DOPE), (4) Reinforcement learning algorithm Soft Actor Critic (SAC), and (5) Residual policy. We train the policy for over 60 iterations, each with two episodes, 1000 steps long.}

\label{fig:real_training_curves}
\end{figure*}

\subsection{Rewards}
For GUAPO, we use a sparse reward when the policy finishes the task (inserts the peg). The policy gets -1 everywhere, and 0 when it finishes the task. For our other learning-based baselines (SAC~\cite{Haarnoja2018sac} and Residual policy~\cite{silver2018residual, Johannink2019residual}), we use a negative L2 norm to the perception estimate of the goal location $\hat{\mu}_{\textit{hole}}$, 0 when it reaches $\hat{\mathcal{S}}_{\textit{u}}$, and 1 when it finishes the task. 

\begin{table*}[t!]
\small
\centering
\caption{Real world peg insertion results out of 30 trials. All learning policies (SAC, Residual, and Guapo) trained for 120 episodes (which takes around 90 minutes).  
The first row indicates percentage success for a full peg insertion. 
The second row depicts the speed of insertion for the trained policies. 
The last two rows indicate the percentage the method enters $\mathcal{S}_{\textit{u}}$ and $\hat{\mathcal{S}}_{\textit{u}}$.}
\begin{tabular}{llllllll}
\multicolumn{1}{l|}{} &  {\begin{tabular}[c]{@{}c@{}}MB- \\Perfect\end{tabular}} & {\begin{tabular}[c]{@{}c@{}}MB- \\DOPE\end{tabular}}& {\begin{tabular}[c]{@{}c@{}}MB-Rand- \\Perfect\end{tabular}}& {\begin{tabular}[c]{@{}c@{}}MB-Rand- \\DOPE\end{tabular}}& SAC~\cite{Haarnoja2018sac} & RESIDUAL~\cite{Johannink2019residual}  & GUAPO (ours)  \\ \cline{1-8} 
\multicolumn{1}{c|}{Success Rate}  & 
\multicolumn{1}{c}{100\%} & 
\multicolumn{1}{c}{0\%} & 

\multicolumn{1}{c}{86.67\%} & 
\multicolumn{1}{c}{26.6\%} &
\multicolumn{1}{c}{0\%} & 
\multicolumn{1}{c}{0\%} & 
\multicolumn{1}{c}{93\%}\\\cline{1-8} 

\multicolumn{1}{l|} {\begin{tabular}[c]{@{}c@{}}Avg. Steps for\\ Task Completion\end{tabular}}  & 
\multicolumn{1}{c}{158.3} & 
\multicolumn{1}{c}{n/a} & 
\multicolumn{1}{c}{554.1} &
\multicolumn{1}{c}{925.4} & 
\multicolumn{1}{c}{n/a} &
\multicolumn{1}{c}{n/a} & 
\multicolumn{1}{c}{469.6}
\\ \cline{1-8}

\multicolumn{1}{c|} {In $\mathcal{S}_{\textit{\textit{u}}}$}  & 
\multicolumn{1}{c}{100\%} & 
\multicolumn{1}{c}{0\%} & 

\multicolumn{1}{c}{100\%} & 
\multicolumn{1}{c}{70.0\%} &
\multicolumn{1}{c}{0\%} & 
\multicolumn{1}{c}{0\%} &

\multicolumn{1}{c}{100\%}\\\cline{1-8}

\multicolumn{1}{c|} {In $\hat{\mathcal{S}}_{\textit{\textit{u}}}$}  & 
\multicolumn{1}{c}{100\%} & 
\multicolumn{1}{c}{100\%} & 

\multicolumn{1}{c}{100\%} & 
\multicolumn{1}{c}{93.3\%} &
\multicolumn{1}{c}{0\%} & 
\multicolumn{1}{c}{100\%}& 

\multicolumn{1}{c}{100\%} \\\cline{1-8} 

\\

\end{tabular}
\label{tab:success}   
\vspace{-0.3cm}
\end{table*}

\section{EXPERIMENTAL DESIGN AND RESULTS}
\label{sec:experimental}

In this section we seek to answer the following questions: 
How does our method compares to our baseline policies, such as, Residual policies, in terms of sample efficiency and task completion? 
And, is the proposed algorithm capable of performing peg insertion on a real robot? 

\subsection{Comparison Methods}

All the different baselines were initialized about 75 cm away from the goal. 
They were all implemented on our real robotics system. 
As such we compare our proposed method to the following:  

\begin{itemize}
 
\item \textbf{MB-Perfect.}
This method consists of a scripted policy under perfect state estimation.

\item \textbf{MB-Rand-Perfect.}
This method uses the same policy as MB-Perfect where we injected random actions, which we sample from a normal distribution with 0 mean and a standard deviation defined by the perception uncertainty from DOPE (which is around 2.5cm to 3 cm). 

\item \textbf{MB-DOPE.} 
This method is similar to MB-Perfect, but instead uses the pose estimator prediction to servo to the hole and accomplish insertion.

\item \textbf{MB-Rand-Dope.}
This method uses the same policy as MB-Dope where we injected random actions, which is sampled in the same way as MB-Rand-Perfect. 

\item \textbf{SAC.} This uses just the policy learned from the RL algorithm, Soft-Actor Critic (SAC), to accomplish the task. 

\item \textbf{Residual.} This method is based off recent residual-learning techniques that combine model-based and reinforcement learning methods~\cite{silver2018residual, Johannink2019residual}.
\end{itemize}

\subsection{Results}

The results comparing the different methods is shown in Table~\ref{tab:success}, this table presents the success rate for insertion 
as well as the average number of steps needed for completion (a step is equivalent to 50 milliseconds of following the same robot command, as our policy is running at 20 Hz), 
and the percentage that the end-effector ends up in the $\mathcal{S}_{\textit{u}}$ and $\hat{\mathcal{S}}_{\textit{u}}$ regions over 30 trials. 
We also present training iteration performance (task success and steps to completion) for the different methods in 
Figure~\ref{fig:real_training_curves}.

MB-Perfect is able to insert 100\% of the time, as it has perfect knowledge of the state, and can be seen as an oracle. We can see that taking random actions with MB-Rand-Perfect does not degrade excessively the full performance achieved by MB-Perfect. However, when we used DOPE as the perception system, which has around 2.5 to 3.5 cm of noise and error, the performance of MB-DOPE and MB-Rand-DOPE drops drastically. MB-Rand-DOPE performs 26.6\% better than MB-DOPE, as the random actions can help offset the perception error.

In our setup SAC did not achieve any insertion. 
This is due to the low number of samples that SAC trained on, since most success stories of RL in the real world require several orders of magnitude more data \cite{levine2018learning}.
The Residual method also did not achieve any insertions. 
The Residual method often would apply large actions far away from the hole opening,
and end up sliding off the box and getting stuck pushing against the side of the box. 
In comparison, GUAPO only turns on the reinforcement learning policy once it is already nearby the region of interest, 
and hence does not suffer from this.
However, Residual was able to reach $\hat{\mathcal{S}}_{\textit{u}}$ 100\% of the time after 120 training episodes, 
while SAC never did.

In comparison, as seen in Fig.~\ref{fig:real_training_curves}, after around 8 training iterations, GUAPO is also able to start inserting into the hole (which is about 12 minute real-world training time). As the policy trains, the average number of steps it takes to insert the peg also decreases. After 120 training episodes (and  90 minutes of training), GUAPO is able to achieve 93\% insertion rate.

\section{RELATED WORK}
\label{sec:related_work}

In robotic manipulation there are two dominating paradigms to perform a task: leveraging model of the environment (model-based method) or leveraging data to learn (learning-based method). The first category of methods relies on a precise description of the task, such as object CAD models, as well as powerful and sophisticated perception systems \cite{schmidt2015dart, wuthrich-iros-2013}. With an accurate model, a well engineered solution can be designed for that particular task \cite{VanWyk2018peg-in-hole, chung2019shallow}, or the model can then be combined with some search algorithm like motion planning \cite{thomas2018assemblyCAD}. This type of model-based approach is limited by the ingenuity of the roboticist, and could lead to irrecoverable failure if the perception system has un-modeled noise and error. 

On the other hand, learning-based approaches in manipulation \cite{levine2016visumotor, haarnoja2018composable} do not require such detailed description, but rather require access to interaction with the environment, as well as a reward that indicates success. Such binary rewards are easy to describe, but unfortunately they render Reinforcement Learning methods extremely sample-inefficient. Hence many prior works use shaped rewards \cite{ lee2019peg-n-hole}, which requires considerable tuning. Other works use low-dimensional state spaces~\cite{zhu2019dexterous} instead of image inputs, which requires either precise perception systems or specially-designed hardware with sensors. There are some proposed methods that manage to deal directly with the sparse rewards, like automatic curriculum generation \cite{florensa2018goal, florensa2017reverse} or the use of demonstrations \cite{Vecerik2017DDPGfD, Ding2019, Nair2017demos}, but these approaches still require large amounts of interaction with the environment. Furthermore, if the position of the objects in the scene changes or there are new distractors in the scene, these methods need to be fully retrained. On the other hand, our method is extremely sample-efficient with a sparse success reward, and is robust to these variations thanks to the model-based component.

Recent works can also be understood as combining model-based and learning-based approaches. One such method \cite{johannsmeier2019haddadin_manipulation} uses a reinforcement learning algorithm to find the best parameters that describe the behavior of the agent based on a model-based template. The learning is very efficient, but at the cost of an extremely engineered pre-solution that also relies on an accurate perception system.
Another line of work that allows to combine model-based and learning-based methods is Residual Learning \cite{Johannink2019residual, silver2018residual}, where RL is used to learn an additive policy that can potentially fully over-write the original model-based policy and does not require any further structure. Nevertheless, these methods are hard to tune, and hardly preserve any of the benefits of the underlying model-based method once trained. 


The problem of known object pose estimation is a vibrant subject within the robotics and computer vision 
communities \cite{tremblay2018deep,hinterstoisser2012accv:linemod,hodan2017wacv:tless,zakharov2019dpod,xiang2018rss:posecnn,hu2019segmentation,peng2019pvnet,Sundermeyer_2018_ECCV,tekin2018cvpr:objpose}.  
Regressing to keypoints on the object or on a cuboid encompassing the object seems to have become the defacto approach for the problem. 
Keypoints are first detected by a neural network, then P$n$P \cite{lepetit2009ijcv:epnp} is used to predict the pose of the object. 
Peng {\em et al.} \cite{peng2019pvnet} also explored the problem of using uncertainty by leveraging a ransac voting algorithm to find 
regions where a keypoint could be detected. 
This approach differs from ours as they do not directly regress to a keypoint probability map, they regress to a vector voting map, where line intersection is then used to find keypoints. 
Moreover their method does not carry pose uncertainty in the final prediction.     


\section{CONCLUSIONS}

We introduce a novel algorithm, Guided Uncertainty Aware Policy Optimization (GUAPO), that combines the generalization capabilities of model-based methods and the adaptability of learning-based methods. It allows to loosely define the task to perform, by solely providing a coarse model of the objects, and a rough description of the area where some operation needs to be performed. The model-based system leverage this high-level information and accessible state estimation systems to create a funnel around the area of interest. We use the uncertainty estimate provided by the perception system to automatically switch between the model-based policy, and a learning-based policy that can learn from an easy-to-define sparse reward, overcoming the model and estimation errors of the model-based part. We show learning in the real world of a peg insertion task.





\section*{ACKNOWLEDGMENT}

Carlos Florensa and Michelle Lee are grateful to all the robotics team at NVIDIA for providing a great learning environment, and providing constant support. Special thanks to Ankur Handa for helping with the compute infrastructure.

\bibliographystyle{IEEEtran.bst}
\bibliography{fullref}



\addtolength{\textheight}{-12cm}   
\end{document}